\documentclass[lettersize,journal]{IEEEtran}
\usepackage{amsmath,amsfonts}
\usepackage{algorithmic}
\usepackage{algorithm}
\usepackage{array}
\usepackage[caption=false,font=normalsize,labelfont=sf,textfont=sf]{subfig}
\usepackage[pagebackref,breaklinks,colorlinks,citecolor=blue]{hyperref}
\usepackage{textcomp}
\usepackage{stfloats}
\usepackage{url}
\usepackage{verbatim}
\usepackage{graphicx}
\usepackage{cite}

\usepackage{makecell}
\usepackage{multirow, hhline}
\usepackage{booktabs}
\usepackage{fontawesome5}
\usepackage{amssymb} 
\usepackage{pifont}
\usepackage{colortbl}
\usepackage{xcolor}
\definecolor{darkblue}{RGB}{0,150,255}
\definecolor{orange}{rgb}{1,0.57,0}
\definecolor{purple}{rgb}{0.478,0.505,1}
\definecolor{dblue}{RGB}{0,150,255}
\definecolor{dpurple}{rgb}{0.533, 0.543, 0.709}
\definecolor{v1}{RGB}{200, 36, 35}
\definecolor{v2}{RGB}{0, 144, 81}
\definecolor{v3}{RGB}{255, 120, 0}
\definecolor{v4}{RGB}{215, 131, 255}
\definecolor{v5}{RGB}{31, 119, 180}
\begin{document}

\title{Show Me When and Where: Towards Referring Video Object Segmentation in the Wild}

\author{Mingqi Gao, Jinyu Yang,~\IEEEmembership{Member,~IEEE}, Jingnan Luo, Xiantong Zhen, Jungong Han,~\IEEEmembership{Senior Member,~IEEE} Giovanni Montana, Feng Zheng,~\IEEEmembership{Member,~IEEE}

\thanks{Manuscript received April 19, 2021; revised August 16, 2021. (Corresponding author: Feng Zheng).}
\thanks{Mingqi Gao is with Department of Computer Science and Engineering, Southern University of Science and Technology, Shenzhen 518055, China, and also with Warwick Manufacturing Group, University of Warwick, Coventry CV1 7AL, U.K. (e-mail: mingqi.gao@outlook.com).}
\thanks{Jinyu Yang is with Tapall.ai, Shenzhen 518055, China (e-mail: jinyu.yang96@outlook.com).}
\thanks{Jingnan Luo is with Department of Computer Science and Engineering, Southern University of Science and Technology, Shenzhen 518055, China. (e-mail: 12332444@mail.sustech.edu.cn).}
\thanks{Xiantong Zhen is with Central Research Institute, United Imaging Healthcare,
Co., Ltd., Shanghai, 201807, China (e-mail:
zhenxt@gmail.com).}
\thanks{Jungong Han is with Department of Computer Science, University of Sheffield, Sheffield S10 2TN, U.K., and also with University of Warwick, Coventry CV1 7AL, U.K. (e-mail: jungonghan77@gmail.com).}
\thanks{Giovanni Montana is with Warwick Manufacturing Group, University of Warwick, Coventry CV4 7AL, U.K. (e-mail: g.montana@warwick.ac.uk).}
\thanks{Feng Zheng is with Department of Computer Science and Engineering, Southern University of Science and Technology, Shenzhen 518055, China (e-mail: f.zheng@ieee.org).}
}

\markboth{Journal of \LaTeX\ Class Files,~Vol.~14, No.~8, August~2021}%
{Shell \MakeLowercase{\textit{et al.}}: A Sample Article Using IEEEtran.cls for IEEE Journals}


\maketitle

\begin{abstract}
Referring video object segmentation (RVOS) has recently generated great popularity in computer vision due to its widespread applications. Existing RVOS setting contains elaborately trimmed videos, with text-referred objects always appearing in all frames, which however fail to fully reflect the realistic challenges of this task. This simplified setting requires RVOS methods to only predict \emph{where} objects, with no need to show \emph{when} the objects appear. 
In this work, we introduce a new setting towards in-the-wild RVOS. To this end, we collect a new benchmark dataset using Youtube Untrimmed videos for RVOS - YoURVOS, which contains 1,120 in-the-wild videos with 7$\times$ more duration and scenes than existing datasets. Our new benchmark challenges RVOS methods to show not only \emph{where} but also \emph{when} objects appear in videos. 
To set a baseline, we propose Object-level Multimodal TransFormers (OMFormer) to tackle the challenges, which are characterized by encoding object-level multimodal interactions for efficient and global spatial-temporal localisation. 
We demonstrate that previous VOS methods struggle on our YoURVOS benchmark, especially with the increase of target-absent frames, while our OMFormer consistently performs well. Our YoURVOS dataset offers an imperative benchmark, which will push forward the advancement of RVOS methods for practical applications.
Our benchmark and code are available at: {\hypersetup{urlcolor=black}\url{https://github.com/gaomingqi/YoURVOS}}.
\end{abstract}


\section{Introduction}
\label{sec:intro}

\IEEEPARstart{R}EFERRING video object segmentation (RVOS) has recently attracted increasing research attention due to its great potential in a broad range of applications, e.g., video editing and augmented reality~\cite{seo2020urvos,yang2022object}. The task aims to localise an object referred to by texts in videos. Different from conventional video object segmentation (VOS), which requires human-annotated masks or scribbles~\cite{wang2021survey}, RVOS calls for no graphical interactions and specifies the targets more naturally. 

\begin{figure*}[t]
\centering
\includegraphics[width=.98\linewidth]{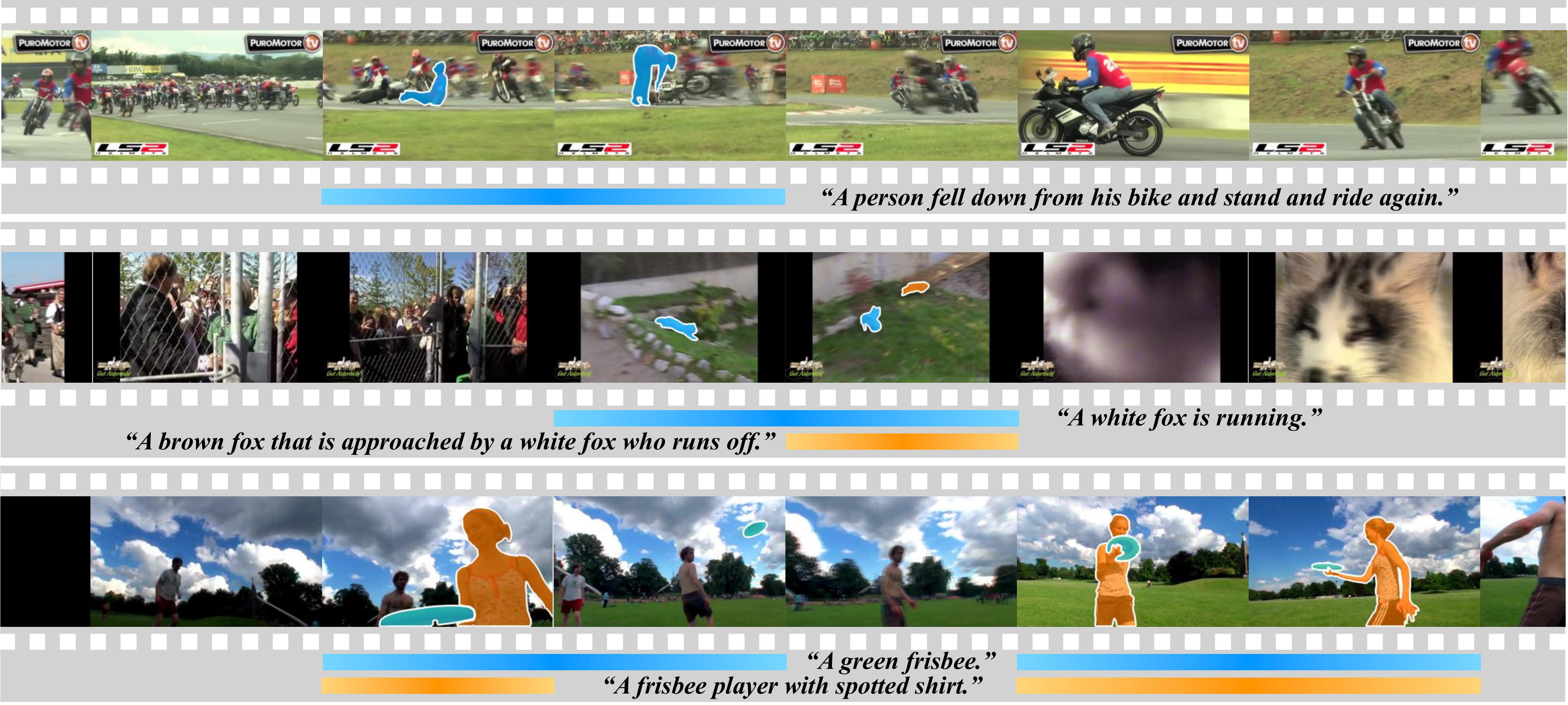}
\caption{Examples from YoURVOS. The text-referred objects are highlighted spatially in blue and orange masks and temporally in thick lines in the same colours. The most unique feature of YoURVOS is untrimmed videos, i.e., all videos are not trimmed to fit the span of any text-referred object. As shown in examples, most objects in YoURVOS appear in one or several segments in a video rather than all frames. With the untrimmed setting, more complex situations are also considered, e.g., broader temporal contexts and multiple scenes. These promote RVOS closer to realistic scenarios and bring several new challenges: \emph{(1)} spatial-temporal joint localisation, \emph{(2)} multimodal analysis on untrimmed videos, and \emph{(3)} long-term video segmentation. }
\label{fig:intro}
\end{figure*}

Previous benchmarks~\cite{gavrilyuk2018actor,khoreva2018video,seo2020urvos,ding2023mevis,yan2024visa,sa2va} contribute significantly to the advances in this field. Nonetheless, these benchmarks are usually constructed from elaborately trimmed videos, where text-referred objects appear on almost all frames. While in realistic scenarios, videos are untrimmed, and the target objects could appear on multiple, disjoint frames in the sequence due to scene changes, camera motion, and other temporal contexts. Therefore, previous benchmarks merely offer a simplified RVOS setting and do not fully represent real challenges of this task. This leads existing RVOS methods to rely on an unreliable assumption that target objects always exist in every frame. In reality, besides \emph{where} the targets are, \emph{when} or in which frame of the video they appear need also to be provided. Considering both enables the RVOS setting to be fully reflective of realistic scenarios. To close the gap, we introduce a new setting towards in-the-wild RVOS task in more realistic, challenging scenarios, in which both \emph{when} and \emph{where} the text-referred objects should be determined in the video.  

To this end, we build up a new dataset - YoURVOS, which collects \underline{Yo}utube \underline{U}ntrimmed videos as the benchmark for \underline{RVOS}. To obtain untrimmed videos, we extend the sampling duration to include more temporal contexts with broad content, scene changes, and diverse scenes. The obtained YoURVOS dataset consists of 1,120 videos with each over 1 minute 30 seconds long on average, which is 7 $\times$ longer than previous RVOS datasets. More importantly, the target objects in YoURVOS are allowed to appear anytime rather than all-time in videos. As shown in Fig.~\ref{fig:intro}, text-referred objects in YoURVOS appear in one (as in the first two examples) or several (as in the last example) video segments. The diverse start/end time and resultant target-irrelevant frames require RVOS methods to study discriminative multimodal interactions, localising text-referred objects in temporal (\emph{when}) and spatial (\emph{where}) domains. 

YoURVOS poses multiple new challenges that are reflective of the realistic RVOS setting. RVOS methods need to handle multimodal interaction over untrimmed videos, spatial-temporal referring segmentation, and efficient training and testing. These have not yet been investigated in existing RVOS methods due to the simplified settings in existing benchmarks, which are usually limited to short videos, with few target-irrelevant distractions. As a result, existing RVOS methods that were usually tuned for short-term temporal contexts exhibit significantly degraded performance on long-term ones. 

Further, we set up a baseline method and propose Object-level multimodal transFormers (OMFormer) for this new, challenging setting. Specifically, we introduce two novel techniques in OMFormer: \emph{(1)} modeling long-term multimodal interaction over object-level features and \emph{(2)} performing spatial-temporal localisation jointly. Unlike existing RVOS methods, which rely on dense features for spatial-temporal modelling, our OMFormer encodes object-level features to model multimodal interactions by capturing spatial-temporal context. By doing so, OMFormer is tailored to effectively handle long-term, untrimmed videos. 

To sum up, we make three major contributions as follows:
\begin{itemize}
\item{We introduce a new setting towards in-the-wild RVOS. We collect a new dataset -  YoURVOS, as a new benchmark for this setting. Compared to previous ones, YoURVOS consists of much longer video sequences with complex contexts, which calls for spatial-temporal joint object localisation. We believe it will foster RVOS studies for more realistic RVOS settings to push the boundary of RVOS towards practical applications.}
\item{We propose Object-level Multimodal transFormers (OMFormer), which is dedicated to tackle long-term untrimmed videos for more realistic RVOS. OMFormer demonstrates its ability to localise objects in both spatial and temporal domains, while requiring much less computation. OMFormer will serve as a baseline model for this challenging setting.}
\item{We conduct extensive experiments and ablation studies. Results show that the performance of existing SoTAs degrades significantly in the new setting, especially in presence of target-irrelevant distractions as commonly encountered in realistic scenarios. These results justify necessity of this new setting and the efficacy of YoURVOS as a benchmark for more realistic, challenging RVOS. OMFormer delivers high performance, showing its effectiveness for RVOS in the wild.}
\end{itemize}

\section{Related Work}
\label{sec:relate}

\subsection{Video Object Segmentation (VOS)} 

VOS aims to segment objects of interest in videos. Existing techniques can be clustered into four categories: unsupervised (or zero-shot), semi-supervised (or one-shot), interactive, and referring VOS. They indicate the targets via saliency (or motion), pixel-level masks, circular scribbles, and language expressions. Here the focus is on the first three, which were initially supported by DAVIS benchmarks (versions 2016~\cite{perazzi2016benchmark} and 2017~\cite{pont20172017}). The former has 50 videos and single-object annotations. The latter was improved with 90 additional videos and multi-object annotations. For further improvement, YouTube-VOS~\cite{xu2018youtube} was built. It has over 4k videos with diverse categories and challenges. DAVIS and YouTube-VOS played a crucial role in VOS development and spawned several classical solutions, e.g., memory-based VOS~\cite{oh2019video,cheng2022xmem}. 

DAVIS and YouTube-VOS only consist of short videos, with an average duration under 5 seconds. As a result, they cannot cover sufficient practical challenges, e.g., temporal association between scenes or frequent reappearance of objects. These problems motivated recent VOS benchmarks: YouMVOS~\cite{wei2022youmvos}, LVOS~\cite{hong2022lvos}, and MOSE~\cite{MOSE}, which are built from longer videos. In particular, YouMVOS consists of 200 videos with an average of 5 minutes long and 75 scenes. LVOS has 220 videos lasting on average 1.29 minutes. MOSE has over 2k videos that are on average 12 seconds long. All these benchmarks are enriched with challenges from longer temporal contexts, encouraging VOS development towards practical applications. 

\subsection{Referring Video Object Segmentation (RVOS)} 

\begin{table*}
  \caption{Comparisons between YoURVOS and previous RVOS datasets. Scene$^{(\mathrm{v})}$ and Dur (v) are the number of scenes (measured by PySceneDetect~\cite{scenedetect}) and the duration per video (measured in seconds, abbreviated as s in the table). Dur (e) and TI are the duration and target-irrelevant rate per expression. Due to sparse annotations, Dur (e) and TI cannot be measured on A2D-Sentences. The largest metrics are presented in \textbf{bold}.
}\label{tab:statistics}\renewcommand{\arraystretch}{1.0}
\centering
\begin{tabular}{|p{0.32\columnwidth}||p{0.12\columnwidth}<{\centering}||p{0.12\columnwidth}<{\centering}||p{0.12\columnwidth}<{\centering}||p{0.12\columnwidth}<{\centering}||p{0.12\columnwidth}<{\centering}||p{0.12\columnwidth}<{\centering}||p{0.12\columnwidth}<{\centering}||p{0.12\columnwidth}<{\centering}|}
\hline
Dataset & Year & Video & Object & Exp & Scene & Dur (e) & Dur (v) & TI\\
\hline
A2D-Sentences~\cite{gavrilyuk2018actor} & 2018 & 3,782 & 4,825 & 6,656 & 1.00 & -- & 4.91s & -- \\
\hline
  J-HMDB-Sentences~\cite{gavrilyuk2018actor} & 2018 & 928 & 928 & 928 & 1.00 & 1.37s & 1.37s & 0.00\% \\
\hline
  Ref-DAVIS~\cite{khoreva2018video} & 2018 & 140 & 255 & 1,644  & 1.00 & 3.11s & 3.45s & 3.21\% \\
\hline
  Ref-YouTube-VOS~\cite{seo2020urvos} & 2020 & \textbf{3,978} & 7,451 & 15,009 & 1.00 & 4.95s & 5.45s & 6.97\% \\
\hline
  MeViS~\cite{ding2023mevis} & 2023 & 2,006 & {8,171} & 28,570 & 1.63 & 10.88s & 13.16s & 15.80\% \\
\hline
  ReVOS~\cite{yan2024visa} & 2024 & 1,042 & \textbf{47,134} & \textbf{35,074} & 1.75 & 10.15s & 13.37s & 16.23\% \\
\hline
\rowcolor{dblue!17!white}
  YoURVOS (Ours) & 2025 & 1,120 & 3,387 & 5,276 & \textbf{11.86} & \textbf{32.08s} & \textbf{92s} & \textbf{72.63\%} \\
\hline
\end{tabular}
\end{table*}

RVOS aims to segment the text-referred objects in videos. RVOS was originally supported by A2D-Sentences~\cite{gavrilyuk2018actor} and JHMDB-Sentences~\cite{gavrilyuk2018actor}, which are built by annotating texts to masked objects from actor and action segmentation datasets: A2D~\cite{xu2015can} and J-HMDB~\cite{jhuang2013towards}. Since A2D and J-HMDB focus on actors, the annotated texts are restricted to action-oriented descriptions. To extend RVOS towards realistic scenes, subsequent benchmarks~\cite{khoreva2018video,seo2020urvos} collect and annotate general objects from VOS datasets. Specifically, Ref-DAVIS~\cite{khoreva2018video} and Ref-YouTube-VOS~\cite{seo2020urvos} are extended from DAVIS-2016/2017 and YouTube-VOS, by annotating language expressions for masked objects. With similar data construction, MeViS~\cite{ding2023mevis}, and ReVOS\cite{yan2024visa} come from diverse video segmentation datasets, encouraging vision-language understanding and reasoning abilities in complex scenes. Tab.~\ref{tab:statistics} summarises key statistics of aforementioned RVOS benchmarks. 

All previous RVOS benchmarks~\cite{gavrilyuk2018actor,khoreva2018video,seo2020urvos,ding2023mevis,yan2024visa} collect videos from public video segmentation datasets, which has two main strengths: \emph{(1)} off-the-shelf masks and \emph{(2)} elaborately selected videos with diverse challenges. Together with the vision-language interaction, these benchmarks are challenging and inspire many innovations. However, they ignored that most source videos have been trimmed to cover target-relevant frames only. As evidenced in Tab.~\ref{tab:statistics}, ALL previous benchmarks have a low Target-Irrelevant (TI) rate, which measures the proportion of frames in a video sequence that do not contain the target. Despite MeViS~\cite{ding2023mevis} and ReVOS~\cite{yan2024visa} containing more target-irrelevant frames than others, the number of these frames is still low, which cannot reflect the real-world scenarios. This simplifies the RVOS setting and drives current studies to assume targets always exist. Therefore, the gap remains between previous benchmarks and practical videos, where real targets appear anytime. In this paper, we build a new benchmark, YoURVOS, from untrimmed videos, where most targets exist in the part of rather than the whole video. Given input videos and texts, we encourage RVOS methods to answer \emph{when} and \emph{where} the text-referred objects are in videos. 

\subsection{Long-term Video Analysis} 

Long-term video analysis has been extensively explored in many realistic tasks, e.g., video understanding~\cite{cohendet2019videomem,wu2021towards,wu2022memvit} and temporal action localisation~\cite{wang2021self,zhu2021enriching,xia2022learning}. Due to coarse granularities, these studies employ frame-level features for long-term contexts. 
At the object level, long-term analysis has been extensively studied in object tracking~\cite{valmadre2018long,fan2019lasot}, where the objective is to maintain the target’s identity over extended temporal spans despite large appearance changes, occlusions, and scene variations. 
Another related line is spatio-temporal visual grounding (STVG)~\cite{zhang2020does,yang2022tubedetr,jin2022embracing}, which aims to localise both the temporal segment and spatial region of a queried instance, often described by natural language, within an untrimmed video, typically at the bounding-box or tube level. However, both tracking and spatio-temporal grounding primarily rely on object-level appearance and motion cues to localise targets with bounding boxes or tubes. In contrast, long-term video object segmentation (VOS) demands pixel-level precision for every frame, requiring the model to simultaneously maintain object-level consistency and fine-grained boundary accuracy across extended durations. This makes the optimisation problem more challenging, as small pixel-wise errors accumulate and erode mask quality over time, limiting the direct applicability of tracking or grounding methods to pixel-level segmentation. 

Long-term VOS presents unique challenges beyond those encountered in tracking or spatio-temporal grounding. Maintaining both object-level consistency and pixel-level accuracy over extended sequences demands substantial computational and memory resources, which quickly become prohibitive in long videos. Consequently, current approaches to long-term video object segmentation (VOS), primarily explored in the semi-supervised setting, focus on designing efficient memory and feature management strategies to balance performance and resource consumption. For example, Liang et al.~\cite{liang2020video} proposed an adaptive memory update mechanism to selectively retain informative frames for long-term VOS. Similarly, XMem~\cite{cheng2022xmem} and Cutie~\cite{cheng2024putting} model temporally local and global dependencies among fine-grained features, achieving improved accuracy while controlling memory footprint. Recently, the SAM-2~\cite{ravi2024sam} architecture has demonstrated remarkable VOS performance thanks to its large-scale pretraining, sparking increased interest in exploring long-term video segmentation built upon SAM-2. However, existing efforts predominantly focus on informative feature management to address computational challenges. For instance, SAM-2-Long~\cite{ding2024sam2long} employs a tree-structured memory mechanism to efficiently store and retrieve relevant information over long sequences, while SAMURAI~\cite{yang2024samurai} leverages motion-based strategies to select informative features dynamically. Nonetheless, these methods still face significant challenges in simultaneously preserving target-level consistency and fine-grained pixel accuracy under resource constraints, motivating further research towards more effective and efficient solutions.

Upon the challenges in long-term video object segmentation, referring video object segmentation (RVOS) introduces even higher demands, yet remains largely unexplored in previous literature. Existing RVOS methods predominantly rely on DETR-based frameworks~\cite{carion2020end,ZhuSLLWD21} to model spatial-temporal relations over dense features, typically focusing on trimmed video clips; however, these approaches struggle to scale effectively to untrimmed, long videos due to computational complexity and the presence of abundant target-irrelevant content. More recently, multi-modal large language model (MLLM) based approaches have emerged~\cite{yan2024visa,bai2024one}, but they face similar limitations due to the high computational cost of dense feature interactions. When dealing with untrimmed, long videos, these methods cannot effectively leverage information from all frames and thus resort to processing only a limited number of key frames within MLLMs. The remaining frames are then processed via semi-supervised VOS methods that propagate masks from key frames, which inevitably leads to partial loss of text-related semantics across the video.

Since existing methods focus heavily on dense feature interactions and their fusion with language, they suffer from prohibitive memory usage on untrimmed videos, making full-frame interaction infeasible. Moreover, traditional trimmed datasets do not adequately reflect these long-term interaction challenges. To better address this gap, we propose YoURVOS, a benchmark and framework designed for long-term referring video object segmentation in untrimmed videos.
As a baseline, we explore reducing computational costs by limiting interactions to the object-level features during the multimodal fusion stage. This approach significantly lowers resource requirements while maintaining effective target representation. We thus develop a simple yet effective baseline, Object-level Multimodal transFormers (OMFormer), to take the initiative towards these goals.

\section{The New Benchmark for In-the-wild RVOS}
\label{sec:dataset}

\begin{figure}[t]
\centering
\includegraphics[width=\linewidth]{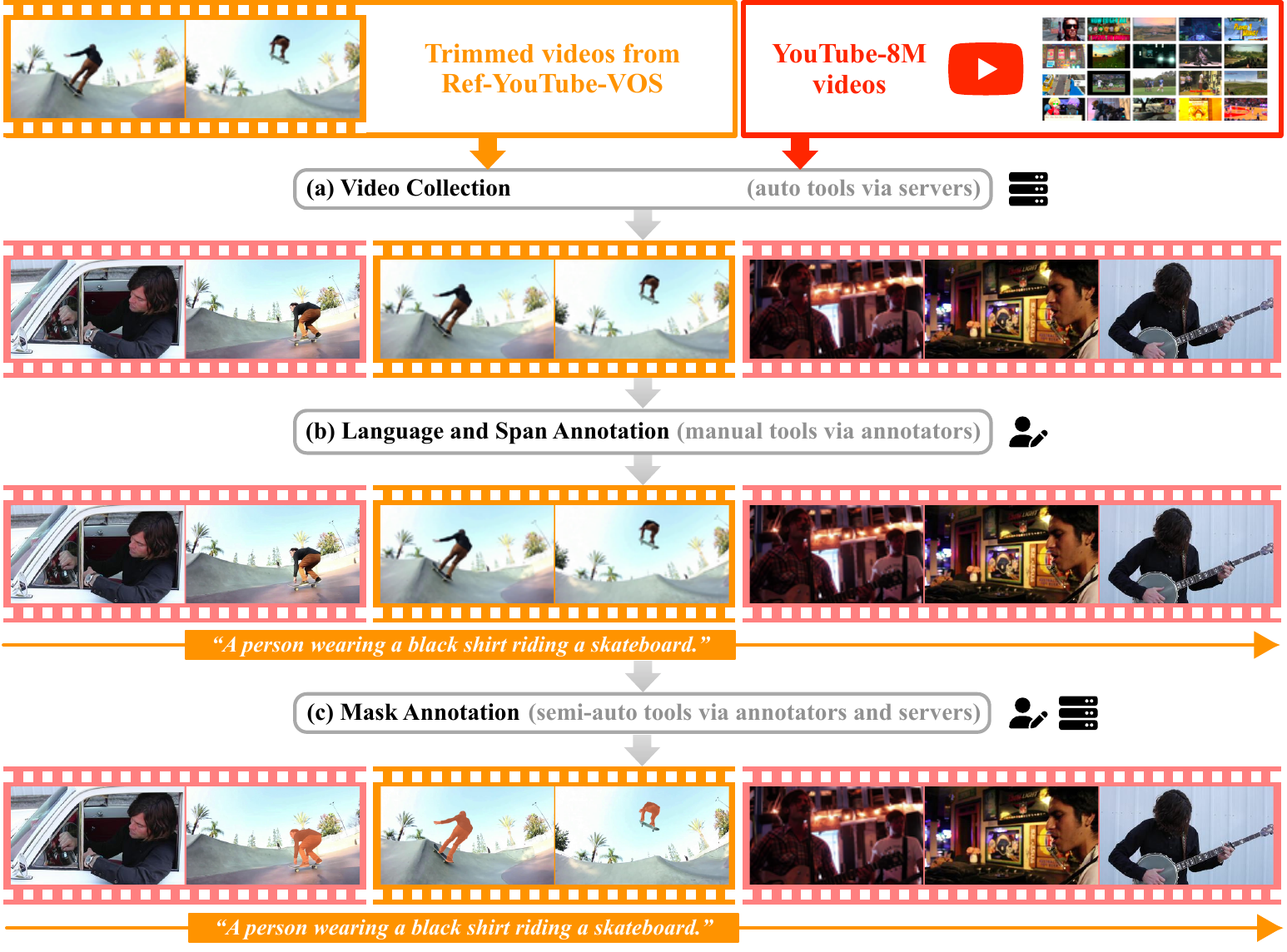}
   \caption{Dataset construction pipeline. (a) For Ref-YouTube-VOS videos (orange film), we retrieve their untrimmed sources from YouTube-8M~\cite{abu2016youtube} to pad before and after (red films); (b) We select target objects from collected videos and annotate corresponding language descriptions and spans (orange thick line); (c) We annotate masks (orange masks) for the target objects. From the example above, it is also clear that YoURVOS videos consider much more complex contexts and scenes than trimmed ones. }
\label{fig:dataset construction}
\end{figure}

Towards RVOS in the wild, we introduce a new setting by building a dataset - YoURVOS with untrimmed videos, which as a new benchmark sets realistic challenges. This section will introduce data construction, statistics, challenges, and evaluation metrics. 

\subsection{Video Collection} Fig.~\ref{fig:dataset construction} outlines our dataset construction. At first, we collect untrimmed videos based on Ref-YouTube-VOS~\cite{seo2020urvos}. We choose Ref-YouTube-VOS since it covers diverse categories and challenges. As discussed in Sec.~\ref{sec:relate}, Ref-YouTube-VOS videos are collected via elaborate trimming. Therefore, retrieving their untrimmed sources is a natural choice for our collection. 

For Ref-YouTube-VOS videos, we do the collection in four steps: \emph{(1)} Retrieve their sources from YouTube-8M~\cite{abu2016youtube} (where all Ref-YouTube-VOS videos trimmed from). Specifically, the retrieval is achieved based on frame-level feature similarities between videos. For YouTube-8M, frame-level features have been encoded as vectors with 1024 dims, via Inception~\cite{szegedy2015going} and Principal Component Analysis (PCA). We download them from its official website\footnote{\url{https://research.google.com/youtube8m/download.html}}. To better measure the similarities, we apply the same encoding pipeline to Ref-YouTube-VOS videos; \emph{(2)} Filter out less challenging sources with fewer scene changes, which is measured by PySceneDetect~\cite{scenedetect}; \emph{(3)} Pad Ref-YouTube-VOS videos with corresponding sources; Specifically, we pad each video with $l_{\mathrm{b}}$ and $l_{\mathrm{a}}$ seconds of source frames before and after. We set $l_\mathrm{b}+l_\mathrm{a}=90$ and sample them randomly to simulate the real temporal distribution of video objects. \emph{(4)} For the wrong sources (due to ambiguous retrieval results) yet sufficient challenges, we randomly split them into videos that are 90 seconds long and keep the challenging ones. With the above steps, we have collected 1000 training and 120 test videos, with an average length of 90 seconds and 6 FPS. More details about our data collection and video visualisations are available on our project page\footnote{\url{https://github.com/gaomingqi/YoURVOS}}. 

\subsection{Video Annotation} Albeit YoURVOS videos are collected upon Ref-YouTube-VOS~\cite{seo2020urvos}, we cannot inherit their text and mask annotations since 
temporal context changes drastically after extending video length from 3 to 90 seconds. Most objects referred to by previous text annotations are ambiguous in untrimmed videos and thus pose difficulties in annotating masks. Therefore, we employed a team with 30 annotators to gear our videos with language/mask annotations from scratch. 

As shown in Fig.~\ref{fig:dataset construction}, YoURVOS is built
in two stages: \emph{(1)} Language and Span annotation and \emph{(2)} Mask annotation. In the first stage, annotators select objects from untrimmed videos and write discriminative texts to specify them. In addition, time spans for these objects are required since most of them only appear within a short moment. For high-quality and efficient annotation, we develop a web-based tool based on VIA~\cite{dutta2019via} to gear annotators with user-friendly UIs to add/verify texts and spans. In the second stage, annotators are provided with videos, texts, and spans to add masks in a semi-automatic way. For each object, annotators manually label a few keyframes and propagate annotations to the remaining frames. With Track-Anything~\cite{yang2023track}, we build a web-based tool to achieve the annotation and focus annotators on the text-referred objects. In particular, annotators select key frames for interactive image segmentation, while masks for other frames are generated through object tracking. To ensure high-quality annotations, we perform manual verification where annotators carefully inspect and refine the masks (especially the auto-generated ones) after the initial annotation process.

\begin{figure*}[tb]
  \centering
  \includegraphics[width=\linewidth]{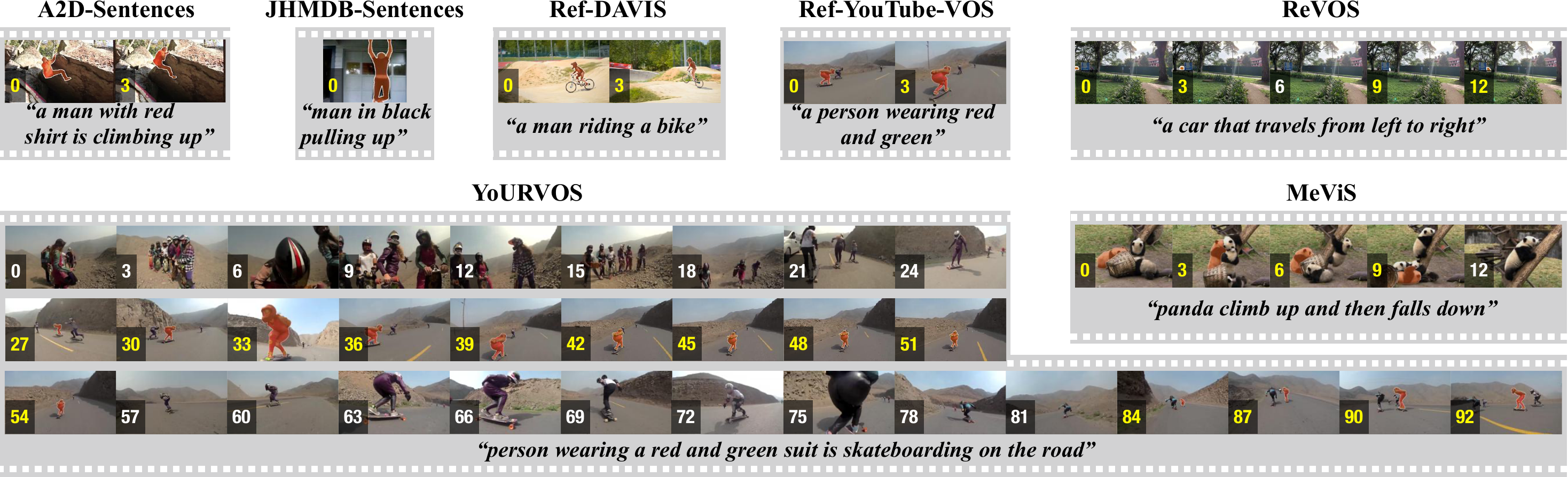}
  \caption{Previous RVOS v.s. RVOS in the wild. Video frames are sampled every 3 seconds and with timestamps left-bottom. Yellow and white numbers indicate target-relevant and -irrelevant frames. Compared to previous benchmarks, YoURVOS is much more challenging and closer to realistic scenarios with long-term videos, target-irrelevant distractions, and multiple scenes, calling for spatial-temporal joint localisation. }
\label{fig:comp}
\end{figure*}

\begin{figure*}[tb]
  \centering
  \includegraphics[width=.99\linewidth]{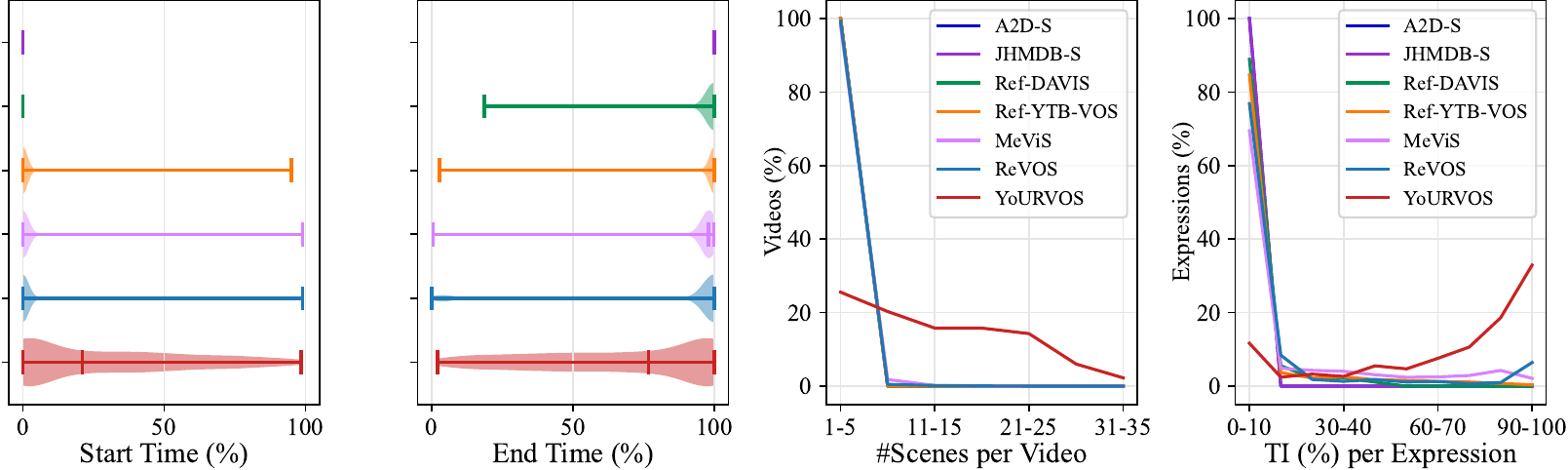}
  \caption{Distributions of RVOS datasets on object span, number of scenes, and TI. For each violin chart summarising start/end times of object appearance, shades depict the probability density. Three markers are minimum/median/maximum data. The reason for plots with less than three markers lies in the over-concentrative data distribution. Violin charts share the same legend as the line charts in this figure. A2D-Sentences~\cite{gavrilyuk2018actor} is not shown in some cases due to its sparse mask annotations. }
\label{fig:violin}
\end{figure*}

\subsection{Dataset Statistics} YoURVOS consists of 1,120 untrimmed videos and 5,276 texts indicating objects to segment. Tab.~\ref{tab:statistics} shows more statistics and comparisons with previous datasets. Note that we merge statistics of Ref-DAVIS-2016 and 2017~\cite{khoreva2018video} for simplicity. Some statistics of Ref-YouTube-VOS~\cite{seo2020urvos} and MeViS~\cite{ding2023mevis} come from their training sets since they only release training annotations. These would not weaken the comparisons significantly as they are measured on average, and their training sets are over 80\% of the whole videos.

Since YoURVOS is constructed based on Ref-YouTube-VOS~\cite{seo2020urvos}, YoURVOS has fewer videos, objects, and expressions than Ref-YouTube-VOS and more recent MeViS~\cite{ding2023mevis}, ReVOS~\cite{yan2024visa}, and Ref-SAV~\cite{sa2va}. Unlike all previous datasets, whose videos are elaborately trimmed towards target-relevant content and single scenes, we focus on RVOS in the wild. As a result, YoURVOS is featured by complex spatial-temporal context, i.e., broad content, diverse scenes, and target-irrelevant distractions, all of these are quantified in Tab.~\ref{tab:statistics} and visualised in Fig.~\ref{fig:comp}. In particular, YoURVOS is at least seven times superior to previous datasets in the duration and number of scenes. Also, YoURVOS is not trimmed for any targets and thus has more target-irrelevant frames, which are common in realistic scenarios and measured quantitatively by Target-Irrelevant rate (TI rate), the ratio of target-irrelevant frames over all frames.

We provide more detailed statistics in Fig.~\ref{fig:violin} to illustrate the characteristics of the proposed YoURVOS further. Specifically, we compare different datasets in the start/end times of object appearance, number of scenes, and TI. It is clear that almost all previous videos are trimmed to fit the target appearance and exclude irrelevant distractions. This, however, tends to be over simplified compared to practical scenarios, where video objects appear/disappear anytime and surrounded by complex spatial-temporal contexts. In contrast, we identify this problem and build YoURVOS with untrimmed videos and realistic target distributions, further enriching the RVOS setting by exploring both \emph{when} and \emph{where} the targets are in videos. 

\subsection{Challenges} YoURVOS covers highly complex scenarios in practice and therefore poses three new challenges to RVOS: 

\begin{itemize}
\item[$\bullet$] Spatial-temporal joint segmentation. Text-referred objects appear or disappear anytime in YoURVOS videos, calling for joint inference in both temporal (\emph{when}) and spatial (\emph{where}) dimensions. 

\item[$\bullet$] Multimodal understanding in untrimmed videos. YoURVOS covers 7$\times$ more spatial-temporal contexts than previous datasets. The diverse objects, events, and scenes pose enormous distractions for multimodal interactions. 

\item[$\bullet$] Long-term video segmentation. With a rate of 6 FPS, each video in YoURVOS has at least 540 frames, which needs high computation and memory overhead, requiring RVOS methods to be both accurate and efficient. 

\end{itemize}

\subsection{Evaluation Metrics}
\label{sec:data:eval}
Unlike previous RVOS benchmarks, which measure spatial metrics only, YoURVOS calls for spatial-temporal joint localisation and thus evaluates RVOS performance in both domains: 
\begin{equation}
\label{eqn:eval}
\small
\mathcal{J} = \frac{\lvert P_m\cap G_m\rvert}{\lvert P_m\cup G_m\rvert},\ \mathcal{F} = \frac{2P_cR_c}{P_c+R_c},\ \mathrm{tIoU} = \frac{\lvert P_s\cap G_s\rvert}{\lvert P_s\cup G_s\rvert},
\end{equation}
\noindent where $P_m, G_m\in \mathbb{R}^{H\times W}$ are the predicted and ground truth masks. $H$ and $W$ are spatial resolutions. $P_c$ and $R_c$ are precision and recall of mask contours. Therefore, $\mathcal{J}$ and $\mathcal{F}$ measure the regional and contour accuracy in the spatial domain. We adopt their average $\mathcal{J}\& \mathcal{F}$ to indicate the overall performance. Instead of evaluating $\mathcal{J}$, $\mathcal{F}$ on every frame (as previous RVOS), YoURVOS does this only when $P_m$ or $G_m$ is not empty for more dedicated evaluation in the spatial aspect. $\mathrm{tIoU}$ measures the temporal accuracy, where $P_s, G_s\in \mathbb{R}^{T}$ are predicted and ground truth spans. $T$ is the video length. 

\section{A Baseline Model}
\label{sec:baseline}


Under the trimmed contexts in previous RVOS settings~\cite{gavrilyuk2018actor,khoreva2018video,seo2020urvos}, existing RVOS methods are tailored for short videos and predict masks on all frames based on the assumption that target objects always exist. Therefore, they utilise dense features for spatial-temporal association and multimodal analysis without any concern about memory costs and long-term modelling. As we show later in experiments, they are computation-intensive and struggle with untrimmed videos, complex contexts, and target-irrelevant distractions. 

Towards RVOS in the wild, we propose a simple yet effective model, Object-level Multimodal transFormers (OMFormer), to set up a baseline for the introduced, challenging setting. Formally, given an input video $\mathcal{V}=\{v_t\in \mathbb{R}^{3\times H\times W}\}^T_{t=1}$ and a text $\mathcal{W}=\{w_l\}^L_{l=1}$, methods need not only predict target masks $\mathcal{M}=\{m_t\in \mathbb{R}^{H\times W}\}^T_{t=1}$, but also the time of appearance $\mathcal{T}\in \mathbb{R}^T$ for text-referred objects since they could appear in any short segments instead of all frames. Here $H$ and $W$ are spatial dimensions, and $T$ and $L$ are the number of frames and words.

\begin{figure*}[t]
\includegraphics[width=.98\linewidth]{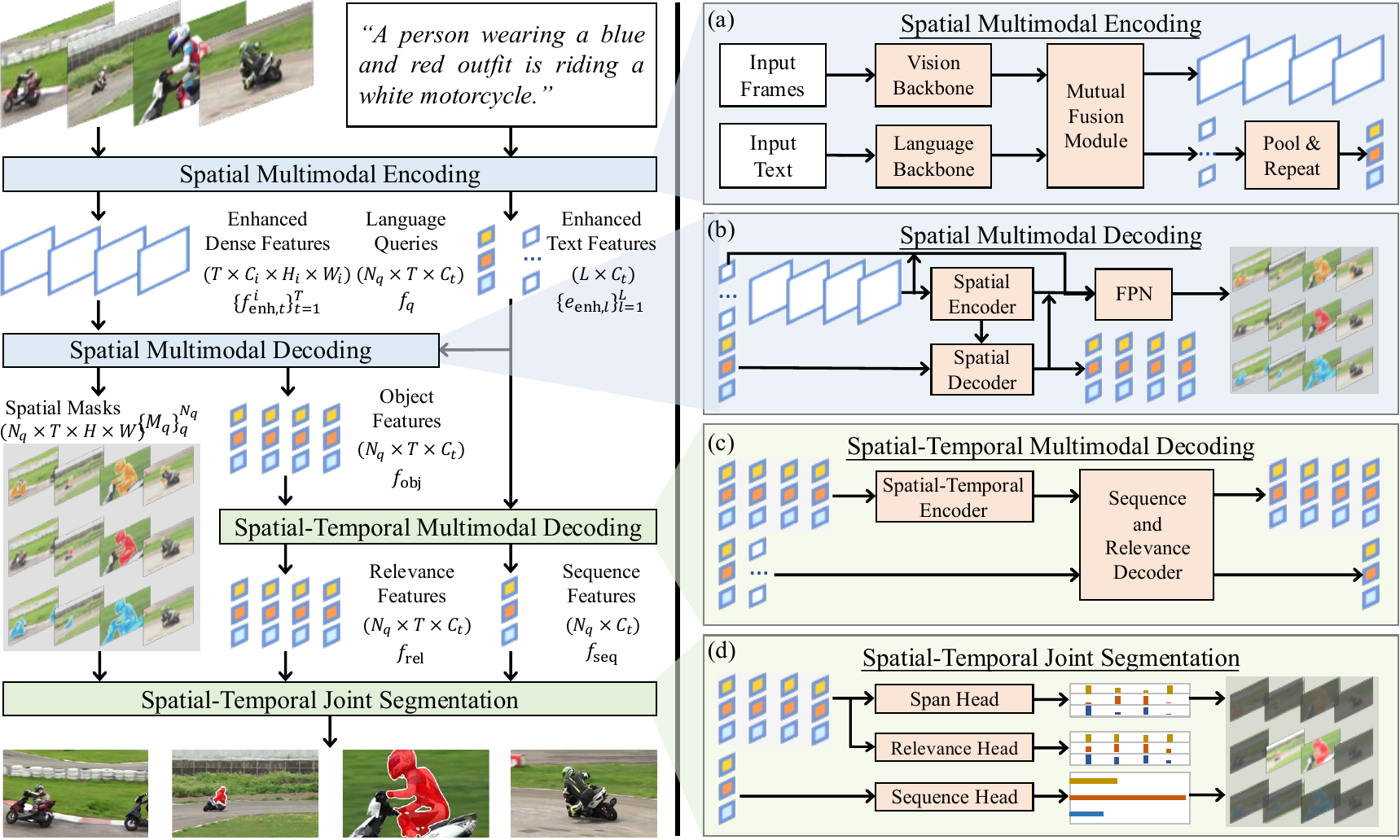}
\caption{Framework of Object-level Multimodal transFormers (OMFormer). \label{fig:diagram}}
\end{figure*}

\subsection{Overview} Fig.~\ref{fig:diagram} diagrams the OMFormer, which performs RVOS in the wild in four stages: \emph{(1)} Spatial Multimodal Encoding, \emph{(2)} Spatial Multimodal Decoding, \emph{(3)} Spatial-Temporal Multimodal Decoding, and \emph{(4)} Spatial-Temporal Joint Segmentation. Instead of dense associations as in previous methods (Fig.~\ref{fig:diagram_compare} (a)), OMFormer encodes fine-grained features for intra-frame operations (stages 1 and 2) and object-level features for inter-frame associations (stage 3, Fig.~\ref{fig:diagram_compare} (b)), making it more efficient to model spatial-temporal and multimodal relations over untrimmed videos. At last, OMFormer localises text-referred targets in spatial/temporal domains, finalising their spatial masks and time of appearance. 




\subsection{Spatial Multimodal Encoding}
As shown in Fig.~\ref{fig:diagram} (a), given a video-text pair ($\mathcal{V}, \mathcal{W}$), OMFormer encodes $v_t$ into $N$ scales of visual dense features: $\{f_{t}^i\in \mathbb{R}^{C_i\times H_i\times W_i}\}^{N}_{i=1}$, $C_i$, $H_i$, and $W_i$ are channel, height, and width dimensions on the $i^{\mathrm{th}}$ scale. Textual features $\{e_l\in \mathbb{R}^{C_t}\}^{L}_{l=1}$ come from $\mathcal{W}$, where $L$ and $C_t$ are the number of word tokens and channels. OMFormer runs mutual cross-attention over $\{f^i_t\}$ and $\{e_l\}$ to improve their correspondence, achieving enhanced dense/text features $\{f^i_{\mathrm{enh},t}\}^{N}_{i=1}$ and $e_{\mathrm{enh},l}$: 
\begin{equation}
\label{eqn:b}
\begin{aligned}
f^{i}_{\mathrm{enh},t} &= \mathrm{Cross\_Attn}(f^i_t,\, e_l),\\
e^{i}_l &= e^{i-1}_l + \mathrm{Cross\_Attn}(e^{i-1}_l,\, f^{i}_{\mathrm{enh},t}),\quad i=1,\ldots,N,\\
e^0_l &= e_l,\quad e_{\mathrm{enh},l}=e^{N}_l,
\end{aligned}
\end{equation}
\noindent where $\mathrm{Cross\_Attn}(q,k)$ denotes the cross-attention with $q$ as query and $k$ as key/value.

\subsection{Spatial Multimodal Decoding} Next, OMFormer integrates enhanced visual dense features $\{f^i_{\mathrm{enh},t}\}^{N}_{i=1}$ into the object level and segments the targets on all frames. Specifically, we follow ReferFormer~\cite{wu2022language} to convert enhanced texts $\{e_{\mathrm{enh},l}\}$ into $N_q$ language queries $f_\mathrm{q}\in \mathbb{R}^{N_q\times C_t}$ (via pool and repeat). For each frame, we aggregate target-relevant contexts from enhanced dense features (with latest scale for high-level semantics) into object-level features:
\begin{equation}
\small
    f_\mathrm{obj} = \mathrm{Cross\_Attn}(f_\mathrm{q}, f^N_{\mathrm{enh},t}).
\end{equation}

As shown in Fig.~\ref{fig:diagram} (b), with dynamic convolution and cross-modal FPN~\cite{wu2022language}, $N_q$ sequences of spatial masks $\mathcal{M}_Q=\{\mathcal{M}_q\}^{N_q}_{q=1}$ are decoded from enhanced features and object features, where $\mathcal{M}_q=\{m^q_{t}\in \mathbb{R}^{H\times W}\}^T_{t=1}$ corresponds to the $q^\mathrm{th}$ object features. 

\subsection{Spatial-Temporal Multimodal Decoding}
\begin{figure*}[t]
\includegraphics[width=.98\linewidth]{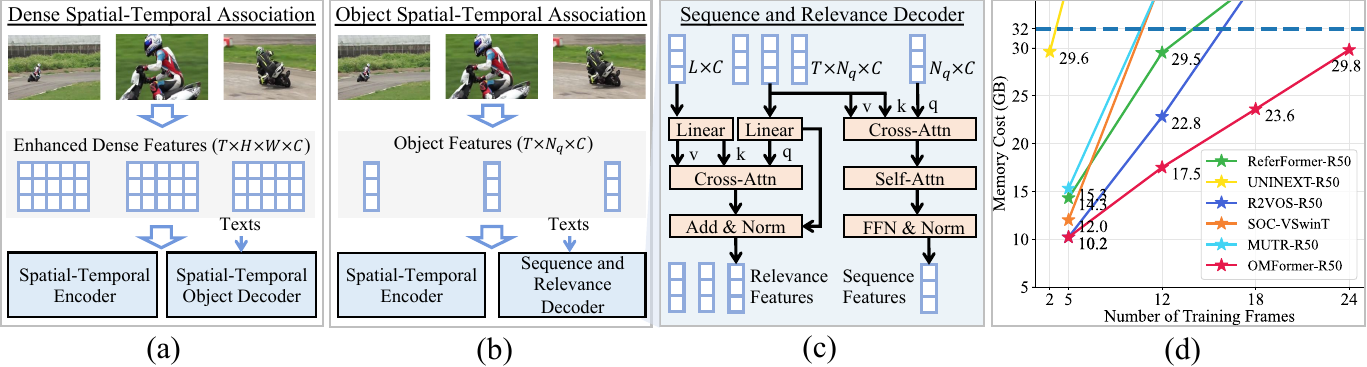}
\caption{Spatial-Temporal Association in (a) Previous RVOS methods and (b) OMFormer. (c) Diagram of the Sequence and Relevance-Aware Decoder; (d) Quantitative comparisons between OMFormer and representative RVOS works in training costs. \label{fig:diagram_compare}}
\end{figure*}

Upon object features, OMFormer builds spatial-temporal associations and aligns them with texts. Specifically, we follow the spatial-temporal encoder in VITA~\cite{heo2022vita} to convert spatial-aware embeddings to spatial-temporal ones. To parse the time of targets' appearance and long-term modelling, we propose a Sequence and Relevance Decoder (SRD). As shown in Fig.~\ref{fig:diagram_compare} (c), it takes word features $e_l$, object features $f_\mathrm{obj}$, and object queries $f_\mathrm{q}$ as inputs and predicts sequence and relevance features $f_\mathrm{seq}\in \mathbb{R}^{N_q\times T\times C_t}$ and $f_\mathrm{rel}\in \mathbb{R}^{N_q\times C_t}$:
\begin{equation}
    f_\mathrm{seq}, f_\mathrm{rel} = \mathrm{SRD}(e_l, f_\mathrm{obj}, f_\mathrm{q}),
\end{equation}
where $f_\mathrm{seq}$ summarises long-term contexts of $N_q$ potential target sequences and their alignments to the referring text. For each sequence, $f_\mathrm{rel}$ measures target relevance on each frame, providing fine-grained clues for subsequent span prediction. 

\subsection{Spatial-Temporal Joint Segmentation} 
In this stage, OMFormer shows when and where text-referred objects appear in videos. Span, Relevance, and Sequence heads are proposed to this end. All consist of an MLP and a sigmoid layers. As shown in Fig.~\ref{fig:diagram} (d), for each sequence, the Span head predicts on each frame the probabilities of targets' start and end appearance ($\{\tau_s, \tau_e\}\in [0,1]^{N_q\times T\times 2}$), and the Sequence head estimates the alignment $c\in [0,1]^{N_q}$ between sequences and texts. The Relevance head predicts frame-level relevance with input texts $r\in [0,1]^{N_q\times T}$, assisting the convergence of the Span head. With $N_q$ sets of predictions, the final outputs are generated from the $j^{\mathrm{th}}$ ($j=\mathrm{argmax}(c)$) spatial mask sequence, within the period of [$\mathrm{argmax}(\tau_{s,j}), \mathrm{argmax}(\tau_{e,j})$]. Therefore, the final prediction $\mathcal{M}=\{m_t\in \mathbb{R}^{H\times W}\}^T_{t=1}$ is formed as:
\begin{equation}
m_t
= \mathbf{1}\!\left\{ \arg\max(\tau_{s,j}) \le t \le \arg\max(\tau_{e,j}) \right\}\, m_t^{\,j}.
\end{equation}
\noindent

Here, $\mathbf{1}\{\cdot\}$ denotes the indicator function; when the condition is false, $m_t=\mathbf{0}\in[0,1]^{H\times W}$, i.e., an empty mask (all zeros, no foreground). 

\subsection{Optimisation}
OMFormer is optimised end-to-end by minimising the losses: 
\begin{equation}
\label{eqn:losses}
\begin{aligned}
\mathcal{L}&=\lambda_\mathrm{cls} \mathcal{L}_\mathrm{cls}(c_j, \hat c) + \lambda_\mathrm{box} \mathcal{L}_\mathrm{box}(b_j, \hat b) + \lambda_\mathrm{mask} \mathcal{L}_\mathrm{mask}(m_j, \hat m)\\&+\lambda_\mathrm{span} \mathcal{L}_\mathrm{span}(\tau_{s,j}, \tau_{e,j}, \hat \tau_s, \hat \tau_e) + \lambda_\mathrm{rel}\mathcal{L}_\mathrm{rel}(r_j,\hat r),
\end{aligned}
\end{equation}
where $\hat c=\{1,...,N_q\}$, $\hat b\in [0,1]^{T\times 4}$, $\hat m\in \{0,1\}^{T\times H\times W}$, $\{\hat \tau_s, \hat \tau_e\}\in [0,1]^{T\times 2}$, and $\hat r\in \{0,1\}^{T}$ are ground-truth labels, boxes, masks (0: background, 1: foreground), spans, and relevance scores (0: target-irrelevant, 1: target-relevant). $\{\hat \tau_s, \hat \tau_e\}$ come from the start and end timestamps (via 1-d Gaussian). $r\in [0,1]^{N_q\times T}$ is predicted from the Relevance Head as an auxiliary term, assisting the Span Head to converge. $j\in [1,N_q]$ indicates predictions with the least cost with ground-truth. We employ focal loss~\cite{lin2017focal} as $\mathcal{L}_\mathrm{cls}$ and $\mathcal{L}_\mathrm{rel}$, L1 and gIoU~\cite{rezatofighi2019generalized} losses as $\mathcal{L}_\mathrm{box}$, DICE~\cite{milletari2016v} and focal losses as $\mathcal{L}_\mathrm{mask}$, Kullback-Leibler divergence loss~\cite{rodriguez2020proposal} as $\mathcal{L}_\mathrm{span}$. $\lambda_\mathrm{cls}$, $\lambda_\mathrm{box}$, $\lambda_\mathrm{mask}$, $\lambda_\mathrm{span}$, and $\lambda_\mathrm{rel}$ are 5, 5, 2, 10, and 5. Like existing RVOS paradigms~\cite{wu2022language}, we pretrain OMFormer on images (w/o $\mathcal{L}_\mathrm{span}$, 12 epochs, learning rate decays by 10 at epoch 8 and 10) and then finetune on YoURVOS (10 epochs, learning rate decays by 10 at epoch 6 and 8). The initial learning rate for the visual backbone is 5e-5, and the rest 1e-4.

\section{Experiments}
\label{sec:experiments}

We perform extensive experiments to verify the effectiveness of the proposed YoURVOS and OMFormer. In particular, We choose 22 representative RVOS methods, including ReferFormer~\cite{wu2022language}, MTTR~\cite{botach2022end}, LBDT~\cite{ding2022language}, UNINEXT~\cite{yan2023universal}, R$^2$-VOS~\cite{li2023r2}, OnlineRefer~\cite{wu2023onlinerefer}, LMPM \cite{ding2023mevis}, SgMg~\cite{miao2023spectrum}, DEVA~\cite{cheng2023tracking}, SOC~\cite{luo2023soc}, MUTR~\cite{yan2023referred}, DsHmp~\cite{he2024decoupling}, VD-IT~\cite{zhu2024exploring}, Video LAVT~\cite{yang2024language}, VISA~\cite{yan2024visa}, HTR~\cite{miao2024temporally}, AL-Ref-SAM 2~\cite{huang2025unleashing}, VideoLISA~\cite{bai2024one}, VRS-HQ~\cite{gong2025devil}, SAMWISE~\cite{cuttano2024samwise}, InstructSeg~\cite{wei2024instructseg}, and ReferDINO~\cite{liang2025referdino} to set up the benchmark. For a comprehensive evaluation, we instantiate these methods with their pre-trained weights on different visual backbones (ResNet~\cite{he2016deep}, ViT~\cite{dosovitskiy2020image}, Swin-Transformer~\cite{liu2021swin}, Video Swin-Transformer~\cite{liu2022video}, ConvNeXt~\cite{liu2022convnet}, Video Diffusion~\cite{wang2023modelscope}, SAM-V1~\cite{kirillov2023segment}, and SAM-V2~\cite{ravi2024sam}) and measure their results on YoURVOS test videos. Although most YoURVOS videos are collected from Ref-YouTube-VOS, there are no overlaps between the YoURVOS test set and Ref-YouTube-VOS training set, where most SoTAs are trained. 
\begin{table*}[tb]
  \caption{Quantitative comparisons on YoURVOS. Backbone (L) denotes the Language Backbone, and Backbone (V) denotes the Vision Backbone. For the Vision Backbone, T, B, L, and H correspond to the Tiny, Big, Large, and Huge model variants. For the Language Backbone, B indicates the number of parameters in billions. The best and the second-best scores in each column are highlighted in \textbf{bold} and \underline{underlined}, respectively.}\label{tab:benchmark}\renewcommand{\arraystretch}{1.0}
  \centering
  \begin{tabular}{|p{0.27\columnwidth} || p{0.1\columnwidth}<{\centering} || p{0.12\columnwidth}<{\centering}
|| p{0.23\columnwidth}<{\centering} || p{0.23\columnwidth}<{\centering} || p{0.12\columnwidth}<{\centering}p{0.12\columnwidth}<{\centering}p{0.12\columnwidth}<{\centering} || p{0.12\columnwidth}<{\centering}|}
\hline
    Method & Year & Venue & Backbone (L) & Backbone (V) & $\mathcal{J}\&\mathcal{F}\uparrow$ & $\mathcal{J}\uparrow$ & $\mathcal{F}\uparrow$ & tIoU$\uparrow$ \\
\hline
    \multirow{6}{*}{ReferFormer~\cite{wu2022language}} & \multirow{6}{*}{2022} & \multirow{6}{*}{CVPR} & \multirow{6}{*}{RoBERTa-base} &  ResNet-50 & 12.0 & 12.1 & 11.9 & 32.2\\
    & & & & ResNet-101 & 22.3 & 22.2 & 22.3 & 33.4\\ 
    & & & &  Swin-T &  22.6 & 22.7 & 22.6 & 34.1\\
    & & & &  Swin-L & 24.9 & 24.6 & 25.2 & 34.4\\
    & & & &  Video-Swin-T &  23.0 & 22.8 & 23.1 & 33.7\\
    & & & &  Video-Swin-B &  24.6 & 24.3 & 24.8 & 34.5\\
\hline
    LBDT~\cite{ding2022language} & 2022 & CVPR & LSTM & ResNet-50 & 14.6 & 14.6 & 14.5 & 32.6\\
\hline
    MTTR~\cite{botach2022end} & 2022 & CVPR & RoBERTa-base & Video-Swin-T &  21.4 & 21.3 & 21.6 & 33.6\\ 
\hline
    \multirow{3}{*}{UNINEXT~\cite{yan2023universal}} & \multirow{3}{*}{2023} & \multirow{3}{*}{CVPR} & \multirow{3}{*}{BERT-base} & ResNet-50 &  23.1 & 22.9 & 23.3 & 32.6\\
    & & & & Conv-L & 24.2 & 23.9 & 24.5 & 32.6\\ 
    & & & & ViT-H &  24.8 & 24.4 & 25.2 & 32.6\\
\hline    
    R$^2$-VOS~\cite{li2023r2} & 2023 & ICCV & RoBERTa-base & ResNet-50 & 24.9 & 25.0 & 24.9 & 35.3\\
\hline
    LMPM~\cite{ding2023mevis} & 2023 & ICCV & RoBERTa-base & Swin-T & 13.0 & 12.8 & 13.3 & 21.9\\
\hline
    DEVA~\cite{cheng2023tracking} & 2023 & ICCV & BERT-base & Swin-L & 21.9 & 21.6 & 22.2 & 33.6\\
\hline    
    \multirow{2}{*}{OnlineRefer~\cite{wu2023onlinerefer}} & \multirow{2}{*}{2023} & \multirow{2}{*}{ICCV} & \multirow{2}{*}{RoBERTa-base} & ResNet-50 & 22.5 & 22.4 & 22.5 & 33.8\\
    & & & &  Swin-L & 25.0 & 24.4 & 25.6 & 34.9\\ 
\hline    
    \multirow{2}{*}{SgMg~\cite{miao2023spectrum}} & \multirow{2}{*}{2023} & \multirow{2}{*}{ICCV} & \multirow{2}{*}{RoBERTa-base} & Video-Swin-T & 24.3 & 24.1 & 24.5 & 34.4\\
    & & & & Video-Swin-B & 25.3 & 25.1 & 25.5 & 34.7\\
\hline
    \multirow{2}{*}{SOC~\cite{luo2023soc}} & \multirow{2}{*}{2023} & \multirow{2}{*}{NeurIPS} & \multirow{2}{*}{RoBERTa-base} & Video-Swin-T & 23.5 & 23.2 & 23.8 & 34.4\\
    & & & & Video-Swin-B & 24.2 & 23.8 & 24.6 & 33.6\\ 
\hline    
    \multirow{5}{*}{MUTR~\cite{yan2023referred}} & \multirow{5}{*}{2024} & \multirow{5}{*}{AAAI} & \multirow{5}{*}{RoBERTa-base} & ResNet-50 & 22.4 & 22.3 & 22.6 & 33.3\\
    & & & & ResNet-101 & 23.3 & 23.1 & 23.4 & 33.7\\
    & & & & Swin-L & 26.2 & 25.9 & 26.5 & 35.1\\
    & & & & Video-Swin-T & 23.2 & 23.1 & 23.4 & 33.5\\
    & & & & Video-Swin-B & 25.7 & 25.5 & 26.0 & 34.6\\
\hline
    DsHmp~\cite{he2024decoupling} & 2024 & CVPR & {RoBERTa-base} & Video-Swin-T & 21.0 & 20.7 & 21.3 & 33.5\\
\hline
VD-IT~\cite{zhu2024exploring} & 2024 & ECCV & RoBERTa-base & Video Diffusion & 25.2 & 25.0 & 25.4 & 34.4\\
\hline
    \multirow{3}{*}{Video LAVT~\cite{yang2024language}} & \multirow{3}{*}{2024} & \multirow{3}{*}{T-PAMI} & \multirow{3}{*}{BERT-base} & Video-Swin-T & 22.4 & 22.3 & 22.5 & 33.2\\
    & & & & Video-Swin-S & 23.9 & 23.4 & 24.5 & 33.6\\ 
    & & & & Video-Swin-B & 24.5 & 23.9 & 25.2 & 33.7\\ 
\hline
VISA~\cite{yan2024visa} & {2024} & {ECCV} & Chat-UniVi-7B & SAM-V1-H & 25.8 & 25.8 & 25.7 & 36.4 \\
\hline
VideoLISA~\cite{bai2024one} & 2024 & NeurIPS & Phi-3.8B & SAM-V1-H & 20.9 & 20.8 & 21.0 & 32.7 \\
\hline
HTR~\cite{miao2024temporally} & 2024 & T-CSVT & RoBERTa & Swin-L & 26.1 & 26.2 & 26.1 & 34.1 \\
\hline
AL-Ref-SAM 2~\cite{huang2025unleashing} & 2025 & AAAI & GPT-4 & SAM-V2 & 26.3 & 26.2 & 26.3 & 34.4 \\
\hline
{VRS-HQ~\cite{gong2025devil}} & {2025} & {CVPR} & Chat-UniVi-7B & SAM-V2-L & 25.9 & 25.9 & 25.8 & 35.4 \\
\hline
{SAMWISE~\cite{cuttano2024samwise}} & {2025} & {CVPR} & RoBERTa-base & SAM-V2-B+ & 22.2 & 22.3 & 22.2 & 33.0\\
\hline
{InstructSeg~\cite{wei2024instructseg}} & {2025} & {ICCV} & Mipha-3B & Swin-B & 24.4 & 24.1 & 24.6 & 32.6 \\
\hline
ReferDINO~\cite{liang2025referdino} & 2025 & ICCV & BERT-base & Swin-B & 27.0 & 26.7 & 27.4 & 35.1\\
\hline
    \rowcolor{dblue!17!white}
    \multirow{2}{*}{{\cellcolor{dblue!17!white}OMFormer}} & \multirow{2}{*}{\cellcolor{dblue!17!white}2025} & \multirow{2}{*}{\cellcolor{dblue!17!white}Ours} & \multirow{2}{*}{\cellcolor{dblue!17!white}RoBERTa-base} & \cellcolor{dblue!17!white}ResNet-50 & \cellcolor{dblue!17!white}\underline{33.7} & \cellcolor{dblue!17!white}\underline{33.6} & \cellcolor{dblue!17!white}\underline{33.8} & \cellcolor{dblue!17!white}\underline{44.9}\\ 
    \cellcolor{dblue!17!white}\vspace{-0.375cm}OMFormer& \cellcolor{dblue!17!white}\vspace{-0.375cm}2025& \cellcolor{dblue!17!white}\vspace{-0.375cm}Ours& 
    \cellcolor{dblue!17!white}\vspace{-0.375cm}RoBERTa-base
    &\cellcolor{dblue!17!white}Video-Swin-T  &\cellcolor{dblue!17!white} \textbf{34.6} &\cellcolor{dblue!17!white} \textbf{34.3} &\cellcolor{dblue!17!white} \textbf{34.9} &\cellcolor{dblue!17!white} \textbf{45.7}\\

\hline
  \end{tabular}
\end{table*}

\begin{figure*}[t]
\includegraphics[width=0.98\linewidth]{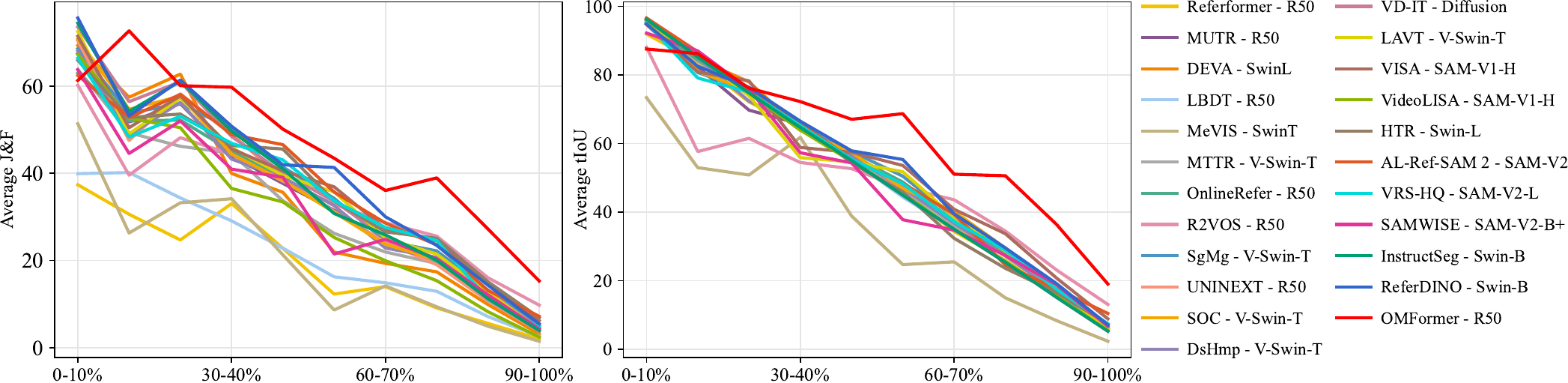}
\caption{Impact of target-irrelevant frames on RVOS results (Left: $\mathcal{J}\&\mathcal{F}$; Right: tIoU). Horizontal axises represent the videos grouped by the TI rate. }
   \label{fig:challenges}
\end{figure*}

\begin{figure*}[t!]
\includegraphics[width=.98\linewidth]{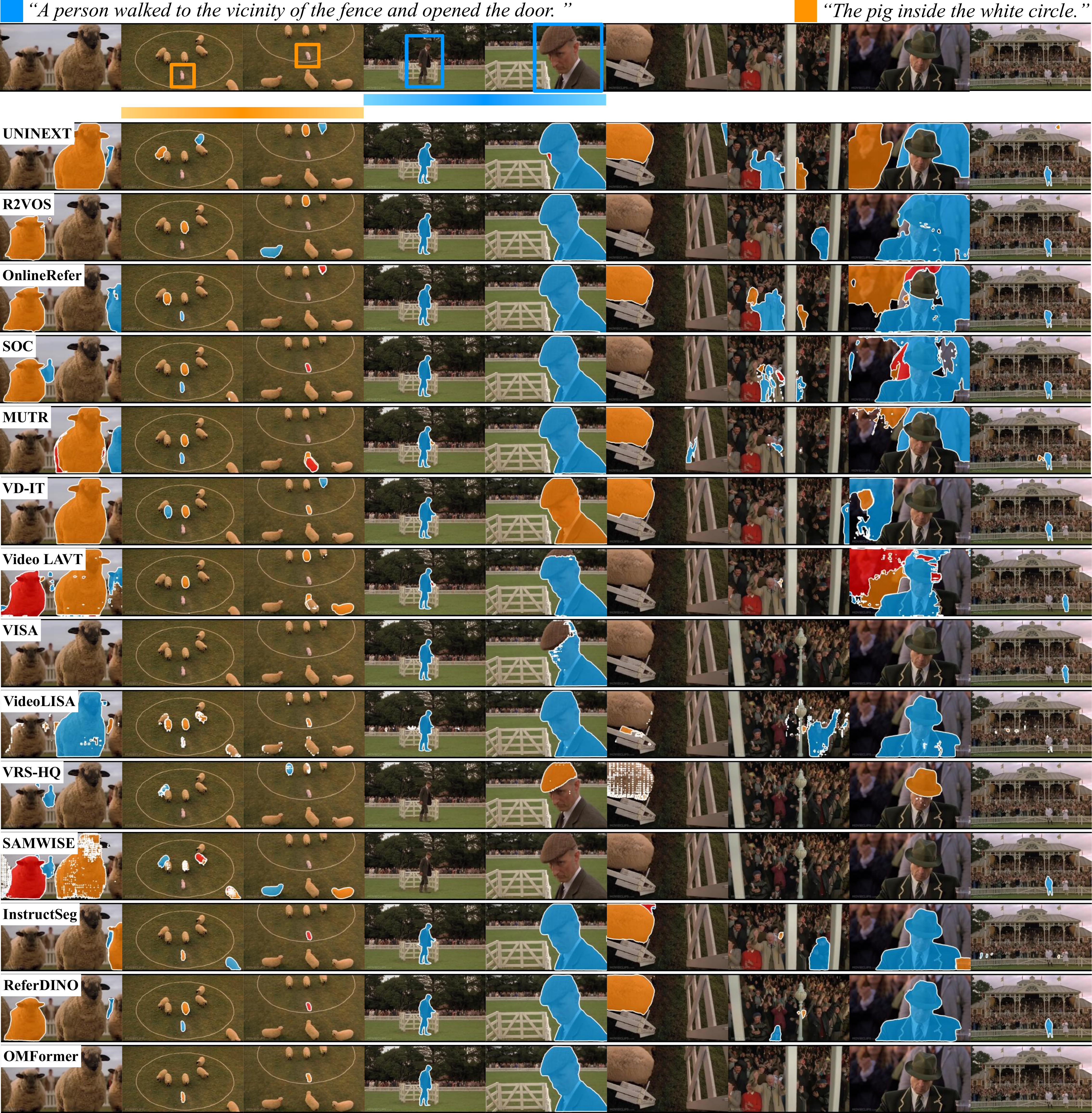}
\caption{Qualitative results on YoURVOS. Blue and orange boxes and lines show text-referred objects spatially and temporally. Red masks are overlap masks. }
   \label{fig:qualitative}
\end{figure*}

\subsection{Quantitative and Qualitative Comparisons}
Tab.~\ref{tab:benchmark} compares quantitative results on YoURVOS. In all metrics, OMFormer outperforms all SoTAs by a large margin, where previous methods listed in Tab.~\ref{tab:benchmark} degrade by an average 18.7 and 31.4 of $\mathcal{J}\&\mathcal{F}$ compared to their records on MeViS and Ref-YouTube-VOS. This is because they were designed under simplified settings, which also shows the necessity of YoURVOS as a new benchmark. Furthermore, we evaluate the methods under different challenge degrees. Specifically, we group YoURVOS test videos by TI and average $\mathcal{J}\&\mathcal{F}$ and tIoU scores over each group. As shown in Fig.~\ref{fig:challenges}, ALL SoTAs degrade significantly with the increase of target-irrelevant content. In tIoU, most degradation is even ``\textbf{LINEAR}'' since they predict masks on all frames. R2VOS is stronger than other SoTA baselines on disappearance but still struggles. In contrast, OMFormer performs better, especially on challenging videos, due to its effective spatial-temporal and long-term multimodal modelling. 

Fig.~\ref{fig:qualitative} visualises results of ours and high-score methods, which are observed to be vulnerable to target-irrelevant content and easy to make false positive predictions. Unlike them, OMFormer localises targets in spatial/temporal domains, showing superior robustness against untrimmed scenes.

\vspace{-3mm}
\subsection{Analysis}

\newcommand{\lowcolorbox}[2]{%
  \makebox[0pt][l]{%
    \colorbox{#1}{\rule[-0.5ex]{\widthof{#2}}{1.5ex}}%
  }%
  #2%
}
\begin{table*}[t!]
\caption{Ablations on spatial-temporal associations and span predictions. `D': Dense; `O': Object, `S': Span predictions, $T_\mathrm{train}$: Number of training frames, `Mem': Memory cost (GB) during training. Scores: Average $\mathcal{J}\&\mathcal{F}$ (left) and tIoU (right) over TI-grouped videos. Numbers in brackets: Number of expressions in each group. }
  \label{tab:ablations}\renewcommand{\arraystretch}{1.0}
  \centering
  \begin{tabular}{|p{0.05\columnwidth}<{\centering}p{0.05\columnwidth}<{\centering}||p{0.1\columnwidth}<{\centering} || p{0.1\columnwidth}<{\centering} || p{0.07\columnwidth}<{\centering}||p{0.12\columnwidth}<{\centering}  p{0.12\columnwidth}<{\centering}
|| p{0.12\columnwidth}<{\centering}  p{0.12\columnwidth}<{\centering}
|| p{0.12\columnwidth}<{\centering}  p{0.12\columnwidth}<{\centering}
|| p{0.12\columnwidth}<{\centering}  p{0.12\columnwidth}<{\centering}|}
\hline
     D & O & $T_\mathrm{train}$ & Mem & S & \multicolumn{2}{c||}{Overall (489)} & \multicolumn{2}{c||}{0\%-33\% (88)} & \multicolumn{2}{c||}{33\%-66\% (77)} & \multicolumn{2}{c|}{66\%-100\% (324)} \\
\hline
    \multirow{3}{*}{ \ding{51}} & \multirow{3}{*}{\ding{55}} & 
    5 & 15.3 & \ding{55} & 27.5 & 32.9 & 79.8 & 90.4 & 41.2 & 50.0 & 9.9 & 13.3\\
    & & 12 & 29.2 & \ding{55} & 23.8 & 32.6 & 80.0 & 90.9 & 28.9 & 48.9 & 7.3 & 12.9\\
    & & 18 & \textgreater 32 & -- & -- & -- & -- & -- & -- & -- & -- & --\\
\hline
\multirow{3}{*}{ \ding{51}} & \multirow{3}{*}{\ding{51}} & \multirow{2}{*}{5} & \multirow{2}{*}{16.5} & 
    \ding{55} & 27.3 & 33.2 & 81.1 & 91.0 & 41.4 & 51.1 & 9.2 & 13.2\\
    & & & & \ding{51} & 11.5 & 16.2 & 36.5 & 49.9 & 11.3 & 17.1 & 4.7 & 6.8 \\
\cline{3-13}
    &  & 12 & \textgreater 32 & -- & -- & -- & -- & -- & -- & -- & -- & --\\
\hline
    \cellcolor{dblue!17!white}& \cellcolor{dblue!17!white}& 
    \multirow{2}{*}{5} & \multirow{2}{*}{10.2} & \ding{55} & 24.8 & 32.9 & 68.1 & 91.4 & 36.6 & 49.3 & 10.3 & 13.1\\
    \cellcolor{dblue!17!white}& \cellcolor{dblue!17!white} & 
    & & \ding{51} & 6.6 & 8.9 & 7.6 & 10.2 & 6.6 & 8.9 & 6.3 & 8.6\\
\cline{3-13}
    \cellcolor{dblue!17!white}& \cellcolor{dblue!17!white}& \multirow{2}{*}{12} & \multirow{2}{*}{17.5} & \ding{55} & 24.6 & 32.9 & 67.8 & 91.2 & 36.0 & 49.5 & 10.1 & 13.1\\
    \cellcolor{dblue!17!white}& \cellcolor{dblue!17!white}& & & \ding{51} & 16.3 & 22.6 & 30.2 & 43.4 & 26.7 & 37.5 & 10.0 & 13.6\\
\cline{3-13}
    \cellcolor{dblue!17!white}& \cellcolor{dblue!17!white}& \multirow{2}{*}{18} & \multirow{2}{*}{23.6} & \ding{55} & 24.2 & 32.7 & 67.4 & 91.4 & 35.2 & 49.0 & 9.7 & 12.8\\
    
    \cellcolor{dblue!17!white}& \cellcolor{dblue!17!white}&  &  & \ding{51} & 28.1 & 37.0 & 50.0 & 66.7 & 37.2 & 51.6 & 19.8 & 26.4\\
\cline{3-13}

     \cellcolor{dblue!17!white} & \cellcolor{dblue!17!white} & \cellcolor{dblue!17!white} & \cellcolor{dblue!17!white} & \ding{55} & 23.9 & 33.0 & 65.7 & 91.4 & 35.1 & 49.8 & 9.8 & 13.2\\

    \rowcolor{dblue!17!white}
    \multirow{-8}{*}{\cellcolor{dblue!17!white}\strut\ding{55}}& \multirow{-8}{*}{\cellcolor{dblue!17!white}\strut\ding{51}}& \multirow{-2}{*}{\cellcolor{dblue!17!white}24} & \multirow{-2}{*}{\cellcolor{dblue!17!white}29.8} & \ding{51} & \textbf{33.7} & \textbf{44.9} & \textbf{64.5} & \textbf{87.1} & \textbf{44.7} & \textbf{62.1} & \textbf{22.7} & \textbf{29.3} \\
\hline
  \end{tabular}
\end{table*}

\paragraph{Object-level Modelling} Tab.~\ref{tab:ablations} compares different levels of spatial-temporal and multimodal modelling. With dense associations only, we infer all frames since span predictions are only available upon object-level associations. As training frames (spatial-temporal contexts) increase, the scores drop and the cost surges significantly, showing that dense associations are costly and fail to model long videos. With object-level associations added, the cost is further enlarged. However, the scores drop since only five frames are insufficient for long-term modelling. In contrast, object-level modelling brings much more efficient computation, enabling long-term analysis and better results in this untrimmed setting. 

\paragraph{Span Predictions} Tab.~\ref{tab:ablations} also verifies the necessity of span predictions. As with previous methods, OMFormers w/o spans segment all frames and outperform the ones w/ spans significantly under lower $T_\mathrm{train}$. However, they cannot handle target-irrelevant distractions and fail to improve with $T_\mathrm{train}$. With broader contexts, more confident spans are derived to enable OMFormer to be robust against untrimmed videos, especially when $T_\mathrm{train}\geq18$. 
\begin{table*}[t!]
\caption{More ablations on $T_\mathrm{train}$. ‘Memory’: Memory cost (GB) during training. Scores: Average $\mathcal{J}\&\mathcal{F}$ (left) and tIoU (right) over TI-grouped videos. Numbers in brackets: Number of expressions in each group. }
\label{tab:train}\renewcommand{\arraystretch}{1.0}
  \centering
  \begin{tabular}{|p{0.1\columnwidth}<{\centering} || p{0.12\columnwidth}<{\centering} || p{0.09\columnwidth}<{\centering}  p{0.09\columnwidth}<{\centering}
|| p{0.09\columnwidth}<{\centering}  p{0.09\columnwidth}<{\centering}
|| p{0.09\columnwidth}<{\centering}  p{0.09\columnwidth}<{\centering}
|| p{0.09\columnwidth}<{\centering}  p{0.09\columnwidth}<{\centering}|}

\hline
     $T_\mathrm{train}$ & Memory & \multicolumn{2}{c||}{Overall (489)} & \multicolumn{2}{c||}{0\%-33\% (88)} & \multicolumn{2}{c||}{34\%-66\% (67)} & \multicolumn{2}{c|}{37\%-100\% (324)} \\

\hline

    20 & 24.0 & 30.6 & 41.0 & 63.3 & 86.4 & 41.5 & 57.5 & 19.1 & 24.8 \\

\hline
    
    22 & 26.7 & 30.3 & 40.7 & 62.9 & 86.3 & 42.8 & 59.3 & 18.4 & 23.9 \\

\hline

    \rowcolor{dblue!17!white}
    \textbf{24} & \textbf{29.8} & \textbf{33.7} & \textbf{44.9} & \textbf{64.5} & \textbf{87.1} & \textbf{44.7} & \textbf{62.1} & \textbf{22.7} & \textbf{29.3} \\

\hline

    26 & 30.9 & 32.5 & 43.2 & 64.6 & 86.6 & 43.4 & 60.2 & 21.1 & 27.3 \\

\hline

    28 & 32.0 & 34.7 & 46.6 & 65.6 & 88.8 & 44.1 & 61.2 & 23.9 & 31.7 \\

\hline

    30 & 34.2 & 34.5 & 45.9 & 66.5 & 89.3 & 45.9 & 63.5 & 23.0 & 29.9 \\

\hline

    32 & 36.0 & 34.9 & 46.1 & 63.3 & 85.2 & 43.9 & 60.6 & 25.0 & 31.9 \\

\hline

    34 & 37.9 & 35.0 & 47.0 & 64.0 & 86.9 & 44.5 & 61.5 & 24.8 & 32.8 \\

\hline
\end{tabular}
\end{table*}

\begin{table}[t!]
\tabcolsep=0.12cm
\caption{Ablations on relevance (Rel) predictions. }
\label{tab:span_sequence}\renewcommand{\arraystretch}{1.0}
  \centering
  \begin{tabular}{|p{0.12\columnwidth}<{\centering}| p{0.07\columnwidth}<{\centering}  p{0.07\columnwidth}<{\centering}
|| p{0.067\columnwidth}<{\centering}  p{0.067\columnwidth}<{\centering}
|| p{0.067\columnwidth}<{\centering}  p{0.079\columnwidth}<{\centering}
|| p{0.073\columnwidth}<{\centering}  p{0.095\columnwidth}<{\centering}|}

\hline
      & \multicolumn{2}{c||}{\scriptsize Overall (489)} & \multicolumn{2}{c||}{\scriptsize0\%-33\% (88)} & \multicolumn{2}{c||}{\scriptsize33\%-66\% (77)} & \multicolumn{2}{c|}{\scriptsize66\%-100\% (324)} \\

\hline
    
    w/o Rel & 32.9 & 44.0 & 64.9 & 87.9 & 43.4 & 60.9 & 21.6 & 28.1 \\

\hline

    \rowcolor{dblue!17!white}
    w/ Rel & \textbf{33.7} & \textbf{44.9} & \textbf{64.5} & \textbf{87.1} & \textbf{44.7} & \textbf{62.1} & \textbf{22.7} & \textbf{29.3} \\

\hline
  \end{tabular}
\end{table}
\vspace{-0.5cm}
\begin{table}[t!]
\scriptsize
\tabcolsep=0.12cm
\caption{Ablations on Sequence and Relevance Decoder.}
\label{tab:sequence_relevance}\renewcommand{\arraystretch}{1.0}
  \centering
  \begin{tabular}{|p{0.12\columnwidth}<{\centering}| p{0.07\columnwidth}<{\centering}  p{0.07\columnwidth}<{\centering}
|| p{0.067\columnwidth}<{\centering}  p{0.067\columnwidth}<{\centering}
|| p{0.067\columnwidth}<{\centering}  p{0.079\columnwidth}<{\centering}
|| p{0.071\columnwidth}<{\centering}  p{0.095\columnwidth}<{\centering}|}

\hline
      & \multicolumn{2}{c||}{\scriptsize Overall (489)} & \multicolumn{2}{c||}{\scriptsize0\%-33\% (88)} & \multicolumn{2}{c||}{\scriptsize33\%-66\% (77)} & \multicolumn{2}{c|}{\scriptsize66\%-100\% (324)} \\

\hline

    w/Decoder & 27.9 & 37.8 & 49.2 & 69.9 & 35.3 & 48.9 & 20.3 & 26.5 \\
    Coupled  & 21.7 & 29.9 & 49.6 & 69.0 & 33.7 & 48.6 & 11.3 & 14.9\\

\hline

    \rowcolor{dblue!17!white}
    Decoupled & \textbf{33.7} & \textbf{44.9} & \textbf{64.5} & \textbf{87.1} & \textbf{44.7} & \textbf{62.1} & \textbf{22.7} & \textbf{29.3} \\

\hline
  \end{tabular}
\end{table}

\paragraph{Relevance Predictions} Tab.~\ref{tab:span_sequence} shows the impact of relevance predictions on the final results. As mentioned in Sec.~\ref{sec:baseline}, they imply the object-text correspondence on each frame and act as an auxiliary term in Eq.~\ref{eqn:losses}. With these fine-grained temporal clues, the span could be predicted more confidently for better results. 

\paragraph{Sequence and Relevance Decoder} Tab.~\ref{tab:sequence_relevance} compares the Sequence and Relevant Decoder (SRD) with other intuitive choices. `No Decoder' indicates identical mapping in Fig.~\ref{fig:diagram_compare} (c). `Coupled' means all inputs are concatenated first and then interacted via self-attention. It is clear that the `Coupled' design troubles the OMFormer in optimisation. The `Decoupled' one aligns texts well with multi-level object embeddings, providing dedicated clues for subsequent predictions. 

\paragraph{Comparisons with Incrementally Fine-Tuned SoTAs}
To demonstrate that better results come from the proposed architecture rather than training data/settings, existing SoTAs are further fine-tuned on the proposed YoURVOS and compared with OMFormer. All training settings of the list methods (e.g., learning rates, number of epochs, number of training frames) are made as in their source literature. As shown in Table~\ref{tab:rebuttal1}, although incremental fine-tuning enables the SoTAs to be compatible with untrimmed videos and brings better performance than training from scratch, OMFormer maintains its superiority due to long-term spatial-temporal multi-modal modelling, validating the effectiveness of the proposed innovations. With incremental fine-tuning, OMformer is observed to perform better, showing that previous RVOS data still contributes to vision-language alignment. Moreover, Table~\ref{tab:rebuttal1} shows that the incremental gains in $\mathcal{J}\&\mathcal{F}$ consistently exceed those in IoU, further indicating that existing methods prioritise spatial segmentation while underutilising temporal correlation.

\paragraph{Comparisons in Complexity}
Although OMFormer is built for more complex tasks (spatial-temporal segmentation) than existing SoTAs (spatial segmentation), only a few complex components are involved. To illustrate this, Table~\ref{tab:rebuttal2} compares the number of parameters (\#Params), inference memory, and FPS (number of Frames Per Second) of different methods. It is evident that with few complex components, OMFormer addresses more challenges. Ablations in this section validates their effectiveness on the performance.

\begin{table}\renewcommand{\arraystretch}{1.0}
\scriptsize
\tabcolsep=0.12cm
  \caption{Quantitative comparisons of OMFormer and existing SoTAs under different training schemes (`Train': same training setting as OMFormer, `Fine-tune': pre-train on other RVOS datasets and fine-tune on YoURVOS), where OMFormer uses Ref-YouTube-VOS, others use more diverse ones. }
  \label{tab:rebuttal1}
  \centering
  \begin{tabular}
  {|p{0.2\columnwidth}|p{0.15\columnwidth}<{\centering}||p{0.078\columnwidth}<{\centering}p{0.078\columnwidth}<{\centering}||p{0.142\columnwidth}<{\centering}p{0.142\columnwidth}<{\centering}|}
  \hline
  \multirow{2}{*}{Method} & \multirow{2}{*}{Backbone} & \multicolumn{2}{c||}{Train} & \multicolumn{2}{c|}{Fine-tune}  \\
  \cline{3-6}
   &  & $\mathcal{J}\&\mathcal{F}$ & IoU & $\mathcal{J}\&\mathcal{F}$ & IoU \\
  \hline
  ReferFormer~\cite{wu2022language} & Resnet-50 & 27.5 & 32.9 & 29.7 (+2.2) & 34.1 (+1.2) \\
  UNINEXT~\cite{yan2023universal} & Resnet-50 & 26.9 & 32.7 & 29.1 (+2.2) & 33.3 (+0.6) \\
  R2VOS~\cite{li2023r2} & Resnet-50 & 27.2 & 33.1 & 29.3 (+2.1) & 33.9 (+0.8) \\
  MUTR~\cite{yan2023referred} & Resnet-50 & 27.3 & 33.2 & 29.2 (+1.9) & 34.0 (+0.8) \\  
  SAMWISE~\cite{cuttano2024samwise} & SAM-V2-B+ & 24.5 & 33.7 & 26.8 (+2.3) & 34.1 (+0.4) \\
  InstructSeg~\cite{wei2024instructseg} & Swin-B & 27.1 & 33.9 & 28.6 (+1.5) & 34.5 (+0.6) \\
  ReferDINO~\cite{liang2025referdino} & Swin-B & 28.1 & 35.2 & 29.9 (+1.8) & 35.7 (+0.5) \\
    \hline
  \rowcolor{dblue!17!white}
  OMFormer & Resnet-50 & \textbf{33.7} & \textbf{44.9} & \textbf{35.1 (+1.4)}  & \textbf{46.2 (+1.3)} \\
\hline
  \end{tabular}
\end{table}

\begin{table}\renewcommand{\arraystretch}{1.0}
  \caption{Comparisons of OMFormer and existing methods with the similar backbone on complexity (during inference) and the quantitative performance on other benchmarks. }
  \tabcolsep=0.12cm
  \scriptsize
  \label{tab:rebuttal2}
  \centering
  \begin{tabular}{|p{0.199\columnwidth}|p{0.12\columnwidth}<{\centering}||p{0.1\columnwidth}<{\centering}p{0.1\columnwidth}<{\centering}||p{0.1\columnwidth}<{\centering}p{0.1\columnwidth}<{\centering}p{0.05\columnwidth}<{\centering}|}
  \hline
  \multirow{2}{*}{Method} & \multirow{2}{*}{Backbone} & R-YTV & MeViS & \#Params & Memory & \multirow{2}{*}{FPS} \\
  \cline{3-4}
   &  & $\mathcal{J}\&\mathcal{F}$ & $\mathcal{J}\&\mathcal{F}$ & (MB) & (GB) &  \\
  \hline
  ReferFormer~\cite{wu2022language} & Resnet-50 & 58.7 & 31.0 & 131.1 & 20.5 & \textbf{3.3} \\
  UNINEXT~\cite{yan2023universal} & Resnet-50 & 61.2 & -- & \textbf{128.5} & \textbf{14.5} & 2.9 \\
  R2VOS~\cite{li2023r2} & Resnet-50 & 60.2 & -- & 132.7 & \textbf{14.5} & 2.8 \\
  MUTR~\cite{yan2023referred} & Resnet-50 & 61.9 & -- & 138.5 & 28.1 & \textbf{3.3} \\
  DsHmp~\cite{he2024decoupling} & Swin-T & \textbf{63.6} & \textbf{46.4} & 272.3 & 39.7 & 2.8 \\
    \hline
  \rowcolor{dblue!17!white}
  OMFormer & Resnet-50 & {62.2} & {45.2} & 132.2 & 20.2 & 3.1 \\
\hline
  \end{tabular}
\end{table}

\vspace{-0.3cm}
\subsection{Limitations and Future Directions}

\begin{figure*}[t]
\includegraphics[width=.98\linewidth]{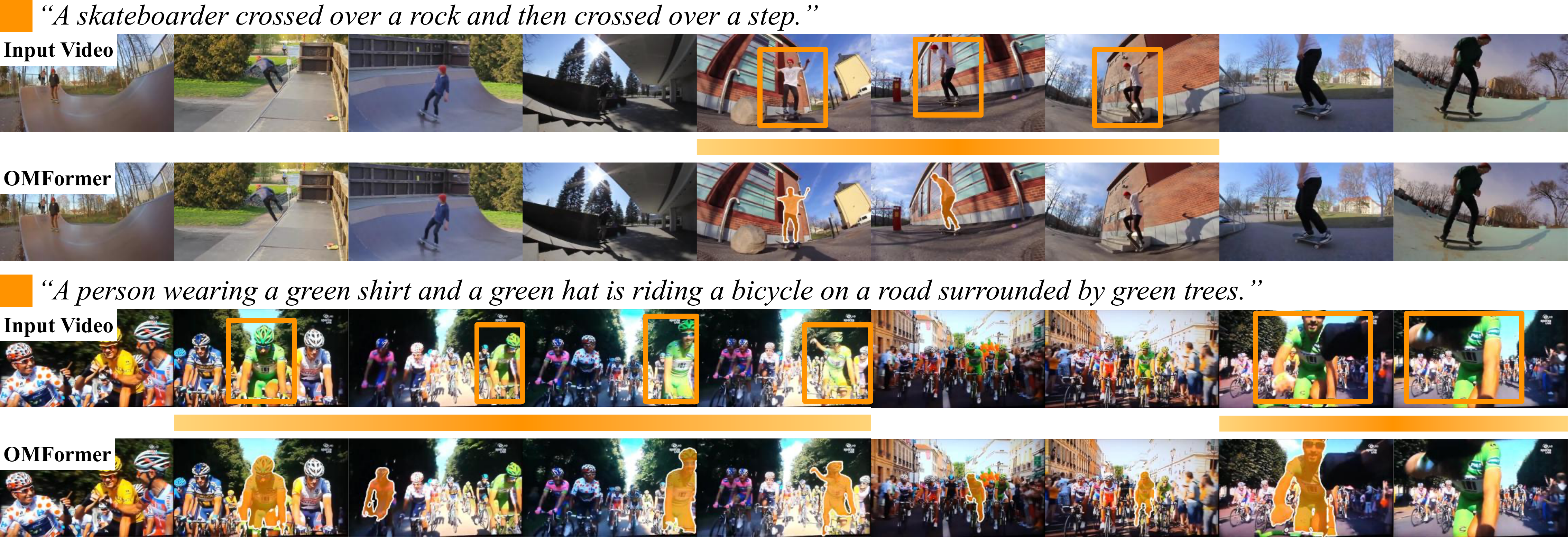}
\centering
   \caption{Failure cases of OMFormer. 
   Orange boxes and lines highlight text-referred objects spatially and temporally. Orange masks are predictions. \label{fig:failure}}
\end{figure*}

Albeit OMFormer is designed towards new challenges in YoURVOS, practical limitations remain which may encourage future research: RVOS with long-term descriptions, multiple distractors, and efficient models. 

\paragraph{RVOS with long-term descriptions.} Most RVOS methods (including OMFormer) prefer data samples with static or short-term temporal descriptions. For example, ``\textit{a skateboarder}'' or ``\textit{a person wearing a white shirt}''. When involving long-term temporal descriptions such as ``\textit{a skateboarder crossed over a rock and then crossed over a step.}'', however, existing RVOS methods usually fail to build long-term multimodal correspondence. As shown in Figure~\ref{fig:failure} (top), OMFormer only corresponds the input video with the first half of the text (``\textit{a skateboarder crossed over a rock}''). With the introduction of YoURVOS based on long videos, we believe the interactions between long-term visual and textual information would be a promising direction for future research. 

\paragraph{Distractors in untrimmed videos} Untrimmed videos introduce substantial target-irrelevant interference; when a video contains many distractor objects and the query includes multiple discriminative cues, OMFormer can still produce erroneous predictions. As illustrated in Fig.~\ref{fig:failure} (bottom), OMFormer sometimes ignores part of the discriminative description, e.g., the clause ``\textit{on a road surrounded by green tree}'', leading to false positive predictions despite the presence of multiple cues. We expect this challenge to be alleviated through more discriminative cross-modal grounding and distractor-aware representation learning in future work.

\paragraph{Efficient Models} OMFormer maximises spatiotemporal segmentation via object-level multimodal interaction with modest computational resources. However, its performance remains constrained by compute and memory. As shown in Tab.~\ref{tab:train}, increasing the number of frames used for spatiotemporal interaction improves accuracy but also increases computational and memory cost. There remains a need for models that support much longer interaction while maintaining computational efficiency. Ideally, such models can cover full videos and even live streams at low computational cost.

\section{Conclusion}
In this paper, we proposed a new benchmark for RVOS in the wild. The dataset, called YoURVOS, has 1,120 videos over 90 seconds long on average, seven times greater than previous RVOS datasets. YoURVOS reflects practical challenges, including scene changes and target-irrelevant distractions. YoURVOS takes a step further towards closing the gap between RVOS and realistic scenarios. In addition, we proposed a baseline method, Object-level Multimodal transFormers (OMFormer), to segment target objects from untrimmed videos. Unlike existing SoTAs, OMFormer captures object features to build spatial-temporal correspondence, achieving global multimodal interactions with less computation. Experimental results validate the necessity of YoURVOS.

\bibliographystyle{IEEEtran}
\bibliography{main}

\end{document}